%% file: main.tex
\documentclass[10pt,twocolumn,letterpaper]{article}

\usepackage{cvpr}              
\usepackage{bm}
\usepackage{epsfig}
\usepackage{graphicx}

\newtheorem{myPro}{Proposition}

\usepackage{multirow}
\usepackage{adjustbox}

\usepackage{amsfonts}       
\usepackage{nicefrac}       
\usepackage{microtype}      
\usepackage{xcolor}         
\usepackage{xspace}
\usepackage{multirow}
\usepackage{adjustbox}
\usepackage{pifont}
\usepackage{bbding}
\usepackage{fontawesome} 
\newcommand{\tabincell}[2]{\begin{tabular}{@{}#1@{}}#2\end{tabular}}
\usepackage{tikz}
\usepackage{soul}

\usepackage[noend]{algpseudocode}
\usepackage{algorithmicx,algorithm}
\usepackage{tabularx}
\usepackage{tikz}
\usepackage{multirow}
\usepackage[mathscr]{euscript}
\usepackage{enumitem}


\definecolor{cvprblue}{rgb}{0.21,0.49,0.74}
\usepackage[pagebackref,breaklinks,colorlinks,allcolors=cvprblue]{hyperref}

\title{Spiking Transformer:\\ Introducing Accurate Addition-Only Spiking Self-Attention for Transformer}

\author{Yufei Guo\thanks{Equal contribution.}, Xiaode Liu\footnote[1]{}, Yuanpei Chen, Weihang Peng, Yuhan Zhang, Zhe Ma\thanks{Corresponding author.}\\
 Intelligent Science \& Technology Academy of CASIC, China\\
{\tt\small yfguo@pku.edu.cn, lxde@pku.edu.cn, mazhe\_thu@163.com}
}

\begin{document}
\maketitle
\input{sec/0_abstract}    
\input{sec/1_intro}

\input{sec/2_formatting}

\input{sec/3_finalcopy}

\section*{Acknowledgment}
This work is supported by grants from the National Natural Science Foundation of China under
contracts No.12202412 and No.12202413.
{
    \small
    \bibliographystyle{ieeenat_fullname}
    \bibliography{main}
}


\end{document}

%% file: sec/0_abstract.tex
\begin{abstract}

Transformers have demonstrated outstanding performance across a wide range of tasks, owing to their self-attention mechanism, but they are highly energy-consuming. Spiking Neural Networks have emerged as a promising energy-efficient alternative to traditional Artificial Neural Networks, leveraging event-driven computation and binary spikes for information transfer. The combination of Transformers' capabilities with the energy efficiency of SNNs offers a compelling opportunity. This paper addresses the challenge of adapting the self-attention mechanism of Transformers to the spiking paradigm by introducing a novel approach: Accurate Addition-Only Spiking Self-Attention (A$^2$OS$^2$A). Unlike existing methods that rely solely on binary spiking neurons for all components of the self-attention mechanism, our approach integrates binary, ReLU, and ternary spiking neurons. This hybrid strategy significantly improves accuracy while preserving non-multiplicative computations. Moreover, our method eliminates the need for softmax and scaling operations. Extensive experiments show that the A$^2$OS$^2$A-based Spiking Transformer outperforms existing SNN-based Transformers on several datasets, even achieving an accuracy of 78.66\% on ImageNet-1K. Our work represents a significant advancement in SNN-based Transformer models, offering a more accurate and efficient solution for real-world applications.

\end{abstract}

%% file: sec/1_intro.tex
\section{Introduction}
\label{sec:intro}

Spiking Neural Networks (SNNs) have gained significant traction as an efficient neural network model, finding applications in diverse areas such as object recognition \cite{li2021free,2021TrainingXiao,Guo_2022_CVPR,guo2022real}, object detection \cite{kim2019spikingyolo,qu2023spiking}, and pose estimation \cite{zou2023eventbased}. These networks utilize binary spike signals for information transfer. A spiking neuron generates a spike, denoted by 1, when its membrane potential surpasses a certain threshold; conversely, it does not spike, represented by 0, when the threshold is not met. This distinctive approach to information processing is notably energy-efficient, as it substitutes the complex multiplications involved in weight and activation interactions with simpler addition operations. Furthermore, SNNs can be effectively implemented using event-driven computations on neuromorphic hardware~\cite{wei2024event,2015Darwin,2015TrueNorth,2018Loihi,2019Towards}. In this setup, the computational units are activated solely upon the occurrence of a spike, conserving energy by remaining inactive in the absence of spikes. Research indicates that SNNs can achieve significantly greater energy savings compared to their Artificial Neural Network (ANN) counterparts~\cite{2015TrueNorth,2018Loihi}.

\begin{figure*}[t]
	\centering
	\includegraphics[width=1.0\textwidth]{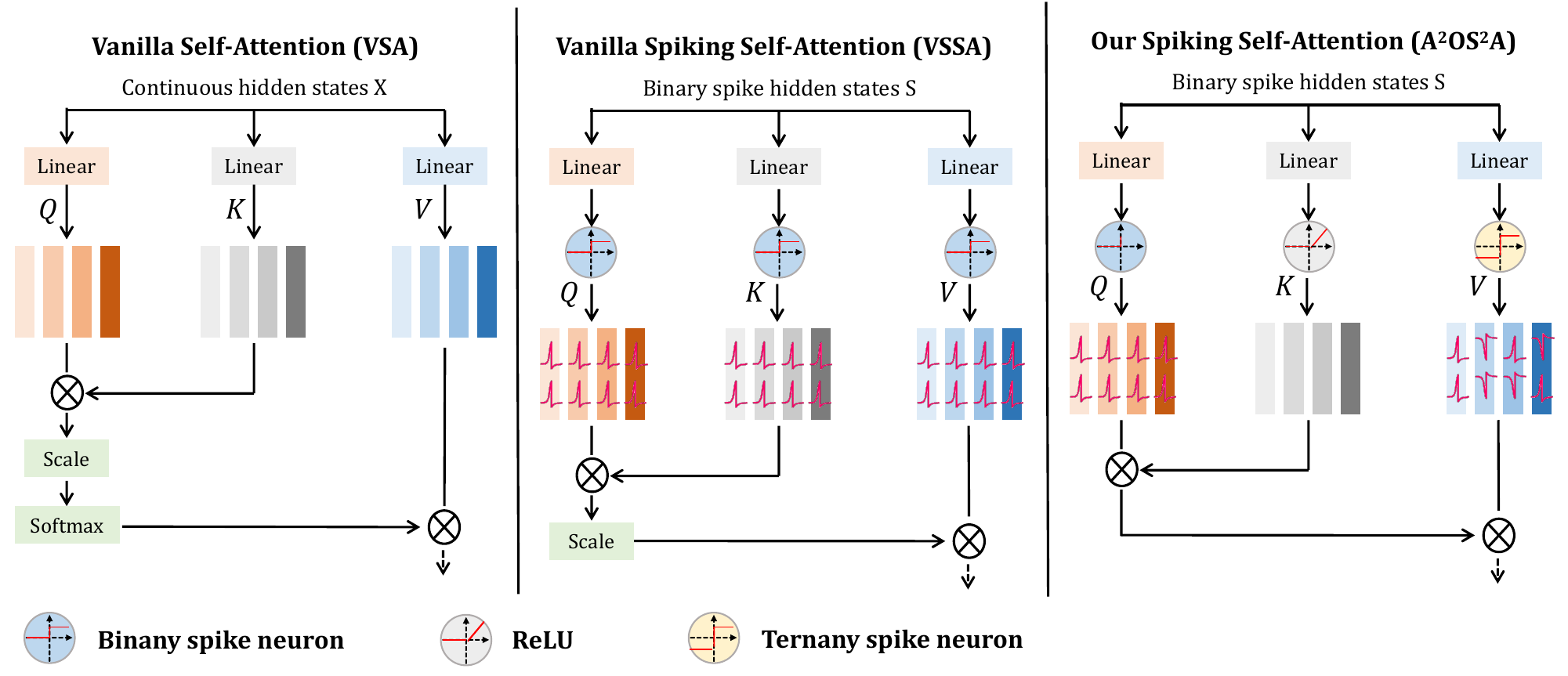} 
	\caption{The difference between our Spiking Self-Attention and the vanilla Spiking Self-Attention. Our Spiking Self-Attention differs significantly from the vanilla Spiking Self-Attention. In the vanilla version, only binary spikes are employed, which can result in considerable information loss. In contrast, our approach integrates a combination of binary spiking neurons, ReLU, and ternary spiking neurons. This hybrid structure effectively reduces information loss, while preserving the advantages of addition-only processing. Moreover, our method eliminates the need for both scaling and the softmax function, streamlining the computation.}
	\label{fig:workflow}
\end{figure*}

Despite the clear advantages of SNNs in terms of energy efficiency, their practical application is often constrained by limited task accuracy. Meanwhile, Transformers have demonstrated exceptional performance across a wide range of tasks, largely due to their self-attention mechanism~\cite{vaswani2017attention,peebles2023scalable,kirillov2023segment}. Combining the strengths of Transformers with the energy efficiency of SNNs presents a promising opportunity.
However, adapting the self-attention mechanism to SNNs is non-trivial. In the vanilla self-attention framework (VSA), three key components are involved: Query ($Q$), Key ($K$), and Value ($V$). As illustrated in Figure~\ref{fig:workflow}(a), the VSA process begins by computing the dot product of the floating-point representations of $Q$ and $K$, resulting in a matrix. This matrix is then normalized using a softmax function, which involves exponential and division operations, to produce an attention map that determines the weighting of $V$.
These operations in the VSA are incompatible with the operational principles of SNNs, which aim to minimize multiplication. Consequently, to implement a Transformer architecture in SNNs, it is essential to develop a novel, efficient self-attention mechanism that avoids multiplication.
Ongoing research efforts are exploring solutions in this domain. Approaches such as Spikformer~\cite{zhou2022spikformer,zhou2024spikformer}, Spikingformer~\cite{zhou2023spikingformer}, and Spike-driven Transformer~\cite{yao2024spike} have focused on transforming $Q$, $K$, and $V$ into spike representations prior to performing matrix operations. This transformation enables the replacement of matrix multiplications with addition, aligning better with the principles of SNNs.

While previous approaches utilize spike-form $Q$, $K$, and $V$ to avoid multiplications, simply introducing spiking neurons to convert $Q$, $K$, and $V$ into spike form prior to performing matrix multiplications does not fully exploit the advantages of the addition-only operation inherent in SNNs. It is unnecessary to convert all of $Q$, $K$, and $V$ into spike form in order to transition from matrix multiplications to matrix additions. However, this method can lead to significant information loss, as demonstrated in Section~\ref{sec:met}.
To address this issue, we propose the Accurate Addition-Only Spiking Self-Attention (A$^2$OS$^2$A). The distinctions between A$^2$OS$^2$A and the vanilla Spiking Self-Attention (VSSA) are illustrated in Figure~\ref{fig:workflow}(b) and Figure~\ref{fig:workflow}(c). Unlike VSSA, which applies the same binary spiking neuron to generate $Q$, $K$, and $V$, our approach employs a binary spiking neuron for $Q$, a ReLU activation for $K$, and a ternary spiking neuron for $V$.
This design allows A$^2$OS$^2$A to retain the multiplication-addition transformation while reducing information loss. Moreover, it eliminates the need for both the scaling and softmax functions, enabling the output of floating-point values similar to the VSA, whereas VSSA produces only integer outputs.
The key contributions of this paper are as follows:

\begin{itemize}
    \item We present a theoretical framework for understanding the information loss associated with vanilla Spiking Self-Attention mechanisms. To the best of our knowledge, this is the first detailed examination of this issue within the context of Spiking Self-Attention, which opens the door for future advancements in SNN-based Transformers.

    \item We introduce A$^2$OS$^2$A, an innovative and efficient approach that incorporates binary spiking neurons, ReLU activations, and ternary spiking neurons, moving beyond the traditional binary-only spiking neuron model. This method reduces information loss while retaining the advantages of addition-only processing in SNNs. Additionally, it eliminates the need for scaling and softmax functions.

    \item Comprehensive experiments demonstrate that our proposed architecture either surpasses or matches the performance of State-of-the-Art (SoTA) SNN-based Transformer models on various datasets. Notably, we achieved an accuracy of 78.66\% on ImageNet-1K, setting a new benchmark in the SNN domain.
\end{itemize}

\section{Related Work}
\label{sec:related_work}

\subsection{Spiking Neural Networks}

SNNs are a class of neural networks that more closely emulate the behavior of biological neurons compared to ANNs. The primary distinction lies in their use of discrete spikes for information transmission, as opposed to the continuous signals employed by ANNs. Foundational work in this area can be traced back to Hodgkin and Huxley~\cite{hodgkin1952currents}, who developed models to simulate the action potentials of biological neurons. Since then, numerous frameworks and models have been introduced to enhance the capabilities and learning efficiency of SNNs. A notable example is the Leaky Integrate-and-Fire (LIF) model~\cite{maass1997networks}, which provides a simple yet effective means of simulating neuronal behavior. More recently, the incorporation of Spike-Timing-Dependent Plasticity (STDP) as a learning rule has enabled SNNs to learn temporal patterns effectively~\cite{caporale2008spike}.

In the past decade, there has been a significant increase in interest regarding the training of SNNs using techniques adapted from ANNs, as well as their application to practical tasks. Two prevalent learning paradigms in current SNN research are the conversion from ANN to SNN (ANN2SNN)~\cite{2019Going, hao2023reducing, hao2023bridging, 2020Deep, li2021free, bu2022optimal, 2020TCL, 2022Optimized, lan2023efficient} and supervised learning~\cite{li2021differentiable, guo2022imloss,Guo2022eccv,guo2023membrane,ren2023spiking,guo2025take,guoenof,guo2024ternary,guo2023joint,zhang2021rectified}. The ANN-SNN conversion method involves initially training an ANN and subsequently converting it into a homogeneous SNN by transferring the trained weights and substituting ReLU neurons with temporal spiking neurons. However, this method often fails for neuromorphic datasets, as ReLU neurons do not adequately capture the complex temporal dynamics required for processing sequential information.
In contrast, supervised learning~\cite{2021Deep, 2018Spatio,zhang2024enhancing,guo2023rmp} employs alternative functions during backpropagation to approximate the firing process, enabling the direct training of SNNs as if they were ANNs. This approach leverages the advantages of gradient-based optimization, achieving impressive performance with just a few time steps, even on large-scale datasets. Furthermore, supervised learning has proven to be effective in handling temporal data, establishing it as a preferred methodology in SNN research.

\subsection{Vision Transformers}

Vision Transformers (ViTs) have recently emerged as a powerful alternative to convolutional neural networks (CNNs) for a wide range of visual tasks, demonstrating remarkable efficacy and flexibility. Introduced by Dosovitskiy et al.~\cite{dosovitskiy2020image}, ViTs leverage the transformer architecture—originally designed for natural language processing—to process images in a novel manner. The architecture innovatively splits images into patches and treats these patches as sequences, enabling the model to capture long-range dependencies and relationships within the visual data, a limitation often encountered in traditional CNNs.
The introduction of ViTs has spurred significant advancements in various computer vision tasks, including image classification~\cite{min2024review}, object detection~\cite{li2022exploring}, and segmentation~\cite{khan2023recent}. These advancements highlight the potential of ViTs to tackle complex visual recognition problems with unprecedented accuracy. Notably, it has been shown that with appropriate training techniques and larger datasets, ViTs can not only match but surpass the performance of state-of-the-art CNNs, even in data-scarce scenarios, presenting a significant advantage in real-world applications~\cite{touvron2021training}.

Furthermore, the adaptation of transformers for visual tasks has inspired numerous variations and improvements across the field. Researchers have explored strategies such as integrating convolutional layers into the architecture~\cite{chen2021crossvit} to retain spatial information more effectively. Other approaches include the use of hierarchical representations~\cite{li2021localvit}, which facilitate multi-scale feature extraction, and the incorporation of attention mechanisms tailored specifically for vision tasks~\cite{caron2021emerging}, improving the model's ability to focus on relevant parts of the input data. These developments underscore the versatility and robustness of the transformer architecture in addressing complex visual challenges, paving the way for further innovations in computer vision.

\subsection{Spiking Neural Networks for Transformers}

The integration of SNNs into transformer architectures represents an exciting and rapidly developing area of research. This field aims to leverage the distinct advantages of both paradigms: the energy-efficient characteristics of SNNs and the robust contextual representation capabilities of transformers. Recent studies~\cite{li2022spikeformer, leroux2023online,wang2025spiking} have explored replacing certain neurons within transformers with spiking neurons, marking a significant step towards this integration.
While these efforts contribute to mitigating the accuracy loss associated with introducing spiking neurons into transformers, they still face challenges in fully realizing the low energy consumption benefits of SNNs. This is primarily due to their reliance on hybrid computing methodologies, which continue to require traditional Multiply-Accumulate (MAC) operations, such as dot products, softmax calculations, and scaling operations.

To address this issue, Spikformer~\cite{zhou2022spikformer, zhou2024spikformer} proposes converting the $Q$, $K$, and $V$ components into spike form before performing matrix multiplications similar to those used in VSA. This innovative approach allows spike matrix multiplications to be transformed into addition operations, eliminating the need for softmax computations. 
However, when considering the residual connections in these architectures, Spikformer~\cite{zhou2022spikformer, zhou2024spikformer} still incorporates non-spiking computations within the ConvBN layers. To overcome this limitation, Spikingformer~\cite{zhou2023spikingformer} and Spike-driven Transformer~\cite{yao2024spike} have restructured the residual connections in standard transformer architectures, placing them before activation functions to ensure that all neurons convey binary spike signals.

The computation of spike-form $Q$, $K$, and $V$ in these models circumvent multiplications, relying solely on additions. However, simply placing spiking neurons in front of $Q$, $K$, and $V$ to convert them into spike form before performing matrix multiplications does not fully exploit the addition-only advantage of SNNs. In this work, we propose a novel, addition-only spiking self-attention mechanism for transformers, aiming to push the boundaries of what can be achieved through the integration of these two technologies.

\section{Preliminary}

\subsection{Spiking Neuron Layer}\label{Sec_spike_layer}
The spiking neuron layer plays a crucial role in integrating both spatial and temporal information, which is then encoded into the membrane potential and subsequently transformed into binary spikes. These spikes drive further computations in subsequent layers of the network. We model the dynamics of the spiking neuron using the LIF model \citep{maass1997networks}. The evolution of the membrane potential and the spike generation mechanism are described by the following equations:
\begin{align}
U[t] &= H[t-1] + X[t], \\ \label{eq_sn_layer}
S[t] &= \operatorname{Hea}(U[t] - V_{\rm th}), \\ 
H[t] &= V_{\rm reset} S[t] + \beta U[t] (1 - S[t]),
\end{align}
where:
 $X[t]$ is the spatial input current at time $t$,
 $U[t]$ is the membrane potential at time $t$, combining the spatial input $X[t]$ and the temporal input $H[t-1]$ from the previous time step,
 $\operatorname{Hea}(\cdot)$ is the Heaviside step function, which outputs 1 if its argument is non-negative, and 0 otherwise,
 $S[t]$ is the spike output at time $t$,
 $V_{\rm th}$ is the spike threshold,
 $V_{\rm reset}$ is the reset potential, and
 $\beta$ is a decay factor controlling the membrane potential’s decay.

When the membrane potential $U[t]$ exceeds a predefined threshold $V_{\rm th}$, the neuron emits a spike, $S[t] = 1$, and the internal state $H[t]$ is reset to $V_{\rm reset}$. If $U[t]$ does not surpass the threshold, the membrane potential decays towards the previous state $H[t-1]$ at a rate governed by $\beta$. To simplify notation, we represent the spiking neuron layer as ${\mathcal{SN}}(\cdot)$, where the input is the membrane potential tensor $U$, and the output is the spike tensor $S$.

\subsection{Vanilla Self-Attention Mechanism (VSA)}\label{sec:Vsa}
The Vanilla Self-Attention (VSA) mechanism enables a model to focus on different parts of the input sequence while constructing its output representations. The self-attention operation can be mathematically expressed as:
\begin{align}
\text{VSA}(Q, K, V) = \text{softmax}\left( \frac{QK^T}{\sqrt{d_k}} \right) V
\end{align}
where \( Q \), \( K \), and \( V \) represent the query, key, and value matrices, respectively, and \( d_k \) is the dimension of the key vectors. The softmax function is applied to normalize the attention scores, ensuring they are non-negative. The resulting weighted sum of the values reflects the importance of each value, based on the similarity between the query and its corresponding key.

The input to the self-attention mechanism consists of a sequence of embeddings \( x_1, x_2, ..., x_n \), which are linearly projected into $Q$, $K$, and $V$ using learned weight matrices:
\begin{align}
Q = XW_Q, \quad K = XW_K, \quad V = XW_V,
\end{align}
where \( W_Q \), \( W_K \), and \( W_V \) are the learned weight matrices corresponding to the $Q$, $K$, and $V$, respectively. 

However, the standard VSA is not directly compatible with SNNs due to the conflict between the floating-point operations required for matrix multiplication of \( Q \), \( K \), and \( V \), and the energy efficiency of SNN computations. What's more, the softmax operation, which involves exponentiation and division, does not align with the computation paradigm of SNNs too.

\begin{figure*}[t]
	\centering
	\includegraphics[width=1.0\textwidth]{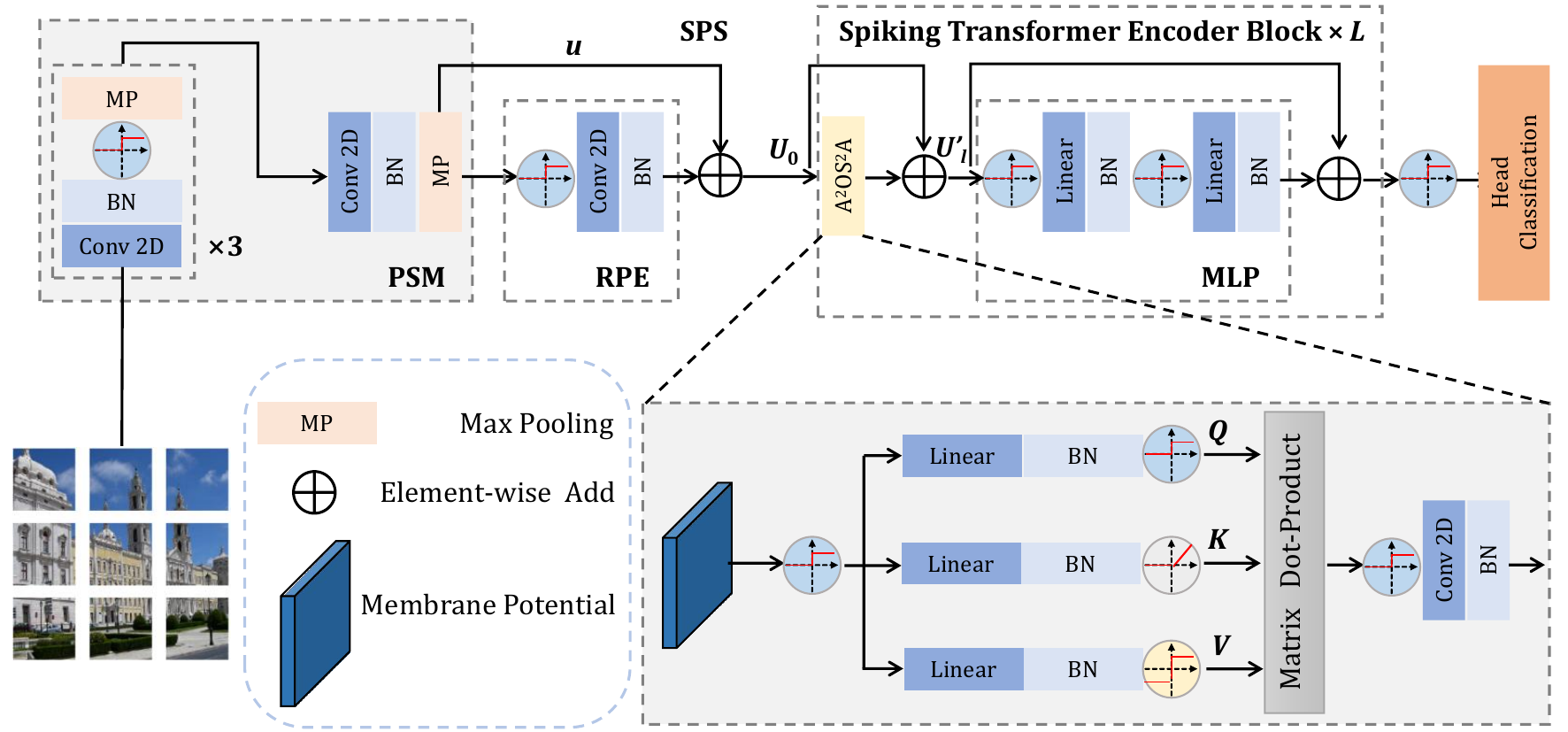} 
	\caption{The overview of Spiking Transformer.}
	\label{fig:overview}
\end{figure*}

\subsection{Vanilla Spiking Self-Attention Mechanism (VSSA)}\label{sec:Vssa}

To address the incompatibility between the VSA and SNNs, the Vanilla Spiking Self-Attention (VSSA) mechanism is proposed in~\cite{zhou2022spikformer,zhou2024spikformer}, which is more suitable for SNNs, as shown in Figure~\ref{fig:workflow}(b). The \( Q \), \( K \), and \( V \) are initially derived through learnable weight matrices. These are then transformed into spiking sequences by applying binary spiking neuron layers:
\begin{align}
Q &= {\mathcal{SN}}_Q({\rm BN}(XW_Q)), \\
K &= {\mathcal{SN}}_K({\rm BN}(XW_K)), \\
V &= {\mathcal{SN}}_V({\rm BN}(XW_V)). 
\end{align}
In this approach, the computation of the attention matrix is performed using binary spike-form queries and keys (which contain only 0s and 1s), thus replacing traditional matrix multiplication with addition-based operations.

To further address the challenges of large values resulting from the matrix multiplication, a scaling factor \( s \) is introduced to regulate the magnitude of the result. The spike-efficient VSSA is formulated as follows:
\begin{align}
\text{VSSA}(Q, K, V) = {\mathcal{SN}}\left( Q ~ K^T ~ V \cdot s \right).
\label{eq:ssa}
\end{align}
In this formulation, the scaling factor \( s \) adjusts the magnitude of the result from the matrix product, while all other operations in VSSA—such as the attention calculation—are performed using addition, in line with the spike-based computation model.

However, simply converting the queries, keys, and values into spike form prior to performing matrix multiplication does not fully leverage the advantages of SNNs, which thrive on addition-only and event-driven computations. It is unnecessary to convert all of \( Q \), \( K \), and \( V \) into binary spike sequences to transition from matrix multiplication to addition. Furthermore, converting all of the values to binary spikes can result in significant information loss. Thus, the challenge remains to strike a balance between maintaining the richness of the data while taking full advantage of the addition operation-based nature of SNNs.

%% file: sec/2_formatting.tex
\section{Method}
\label{sec:met}
\subsection{Overall Architecture}

Figure~\ref{fig:overview} illustrates the architecture of our Spiking Transformer, which comprises four primary components: Spiking Patch Splitting (SPS), Accurate Addition-Only Spiking Self-Attention (A$^2$OS$^2$A), Multi-Layer Perceptron (MLP), and a linear classification head. The design follows the approach outlined in~\cite{yao2024spike}. For the SPS module, we adopt the structure from~\cite{zhou2022spikformer, zhou2023spikingformer}. 
Given a 2D image sequence $ I \in \mathbb{R}^{T \times C \times H \times W} $, the Patch Splitting Module (PSM) performs a linear projection and splits the input into a sequence of $ N $ flattened spike patches with $ D $-dimensional channels, where $ T $ denotes the number of timesteps (images are repeated $ T $ times in the dataset), $ C $ is the number of channels, and $ H $ and $ W $ represent the height and width of the image sequence. Additionally, a convolutional layer generates Relative Position Embeddings (RPE) as described in~\cite{yao2024spike}. Together, the SPS part is written as: 
\begin{align}
    u={\rm{PSM}}\left(I\right), {{I}} \in \mathbb{R}^{T \times C\times H\times W}, x\in \mathbb{R}^{T\times N\times D},
\end{align}
\begin{align}
    s={\mathcal{SN}}(u), s\in \mathbb{R}^{T\times N\times D},
\end{align}
\begin{align}
{\rm{RPE}}={\rm{BN}}(({\rm{Conv2d}}(s))), {\rm{RPE}}\in \mathbb{R}^{T \times N\times D},
\end{align}
\begin{align}
     U_0 = u + {\rm{RPE}}, U_0\in \mathbb{R}^{T \times N\times D},
\end{align}
where $ u $ and $ U_0 $ are the output membrane potential tensors from the PSM and SPS modules, respectively, and $ {\mathcal{SN}}(\cdot) $ represents the spike neuron layer. 
The output $ U_0 $ is then passed to the $L$-block Spiking Transformer encoder, which consists of the A$^2$OS$^2$A and MLP blocks. Residual connections are applied to the membrane potentials in both the A$^2$OS$^2$A and MLP blocks. To avoid multi-bit spike outputs, we adopt the design from~\cite{yao2024spike, zhou2023spikingformer}, ensuring that the values before the convolutional layers are binary, thus allowing the spike and weight matrix multiplication to be simplified to addition operations. 

The A$^2$OS$^2$A mechanism models both local and global information of the input sequence using binary spike $ Q $, full-precision $ K $, and ternary spike $ V $. This approach reduces information loss and eliminates the need for scaling and softmax operations, thereby improving task accuracy. A Global Average Pooling (GAP) is applied to the processed feature from the Spiking Transformer encoder, and the resulting $ D $-dimensional channel is passed to a fully connected Classification Head (CH) for final prediction. The A$^2$OS$^2$A, MLP, and CH components are as follows:
\begin{align}
    S_0={\mathcal{SN}}(U_0), S_0\in \mathbb{R}^{T\times N\times D},
\end{align}
\begin{align}
U^{\prime}_l = {\rm{A^2OS^2A}}(S_{l-1}) + U_{l-1}, U^{\prime}_l\in \mathbb{R}^{T \times N\times D},l=1...L ,
\end{align}
\begin{align}
    S^{\prime}_l={\mathcal{SN}}(U^{\prime}_l), S^{\prime}_l\in \mathbb{R}^{T\times N\times D},
\end{align}
\begin{align}
S_l = {\mathcal{SN}}({\rm{MLP}}(S^{\prime}_l) + U^{\prime}_l)
, S_l\in \mathbb{R}^{T \times N\times D}, l=1...L ,
\end{align}
\begin{align}
     Y = {\rm{CH}}({\rm GAP}(S_L)).
\end{align}

\subsection{Information Loss in Spiking Transformers}

In this section, we analyze the limitations of binary spikes in SNNs with respect to information representation, which motivates the focus of this paper. While binary spikes offer energy efficiency, they inherently suffer from reduced representational capacity, leading to information loss when used for all $ Q $, $ K $, and $ V $ in VSSA, which results in accuracy degradation.

To support this claim, we first perform a theoretical analysis using the concept of entropy. The representational capability $ \mathcal{C}(\mathbf{X}) $ of a set $ \mathbf{X} $ is determined by the maximum entropy of $ \mathbf{X} $, expressed as:
\begin{equation}\label{eq:entroloss}
	\mathcal{C}(\mathbf{X}) = \max \mathcal{H}(\mathbf{X}) = - \sum_{x \in \mathbf{X}} p_{\mathbf{X}}(x) \log p_{\mathbf{X}}(x),
\end{equation}
where $ p_{\mathbf{X}}(x) $ is the probability of a sample $ x $ from $ \mathbf{X} $. We now present the following proposition:

\begin{myPro}
\label{pro:entro1}
    For a set $ \mathbf{X} $, its representational capacity is $ \mathcal{C}(\mathbf{X}) = \max \mathcal{H}(\mathbf{X}) $. When the probability distribution of $ \mathbf{X} $ is uniform, i.e., $ p_{\mathbf{X}}(x) = \frac{1}{N} $, where $ N $ is the total number of samples in $ \mathbf{X} $, the entropy $ \mathcal{H}(\mathbf{X}) $ reaches its maximum value of $ \log(N) $. Hence, we conclude that $ \mathcal{C}(\mathbf{X}) = \log(N) $.
\end{myPro}

Using Proposition~\ref{pro:entro1}, we can evaluate the representational capacity of binary spike layers in SNNs and compare them with real-valued layers in ANNs.
Let $ \mathbf{F}_{\rm B} \in \mathbb{B}^{C \times H \times W} $ denote the binary spike layer of the SNN, and $ \mathbf{F}_{\rm R} \in \mathbb{R}^{C \times H \times W} $ denote the real-valued feature map in the corresponding ANN. For a binary spike output $ s $, it requires 1 bit, and thus the number of possible samples from $ s $ is 2. Therefore, the number of samples for $ \mathbf{F}_{\rm B} $ is $ 2^{(C \times H \times W)} $, leading to:
\begin{equation}
	\mathcal{C}(\mathbf{F}_{\rm B}) = \log\left( 2^{(C \times H \times W)} \right) = C \times H \times W.
\end{equation}
In contrast, a real-valued output requires 32 bits per sample, yielding $ 2^{32} $ possible values. Hence, the representational capacity for real-valued layers is:
\begin{equation}
	\mathcal{C}(\mathbf{F}_{\rm R}) = \log\left( 2^{32 \times (C \times H \times W)} \right) = 32 \times C \times H \times W.
\end{equation}
This highlights the limited representational capacity of the binary spike layer. By transforming the real-valued $ Q_\mathbb{R}, K_\mathbb{R}, V_\mathbb{R} $ into binary $ Q_\mathbb{B}, K_\mathbb{B}, V_\mathbb{B} $, significant information loss occurs, reducing accuracy. Therefore, we propose increasing the bit precision of $ Q $, $ K $, and $ V $ in spiking attention to minimize information loss, while preserving the advantages of multiplication-addition transformations.

\subsection{Accurate Addition-Only Spiking Self-Attention Mechanism (A$^2$OS$^2$A)}

To maintain the integrity of information within spiking attention while leveraging the advantages of multiplication-addition transformations, we propose the Accurate Addition-Only Spiking Self-Attention (A$^2$OS$^2$A). This mechanism employs binary \( Q \), full-precision \( K \), and ternary \( V \) in the self-attention process. Specifically, we utilize a binary spiking neuron after the linear layer to produce \( Q \), a ReLU function for generating \( K \), and a ternary spiking neuron for producing \( V \), as illustrated in Figure~\ref{fig:workflow}(c). The formulations for \( Q, K, V \) in A$^2$OS$^2$A are defined as follows:
\begin{align}
Q &= {\mathcal{SN}}^b_Q(\mathrm{BN}(XW_Q)), \\
K &= \mathrm{ReLU}_K(\mathrm{BN}(XW_K)), \\
V &= {\mathcal{SN}}^t_V(\mathrm{BN}(XW_V)), 
\end{align}
where \({\mathcal{SN}}^b(\cdot)\) denotes the binary spiking neuron as used in standard SNNs, \(\mathrm{ReLU}_K(\cdot)\) is the ReLU activation function typical in ANNs, and \({\mathcal{SN}}^t(\cdot)\) represents a ternary spiking neuron that outputs values in the set \(\{-1, 0, 1\}\).

The evolution of the membrane potential and the spike generation mechanism within \({\mathcal{SN}}^t(\cdot)\) can be described by the following equations:
\begin{align}
U[t] &= H[t-1] + X[t], \\ 
S[t] &= \operatorname{Hea}(|U[t]| - V_{\rm th}), \\ 
H[t] &= V_{\rm reset} S[t] + \beta U[t] (1 - |S[t]|).
\end{align}
The formulation of the A$^2$OS$^2$A is expressed as:
\begin{align}
\text{A$^2$OS$^2$A}(Q, K, V) = {\mathcal{SN}}\left( Q \cdot K^T \cdot V \right).
\label{eq:vssa}
\end{align}
In A$^2$OS$^2$A, since \( Q \) takes values in \(\{0, 1\}\) and \( K \) is in \(\mathbb{R}\), the matrix multiplication \( Q \cdot K^T \) can be transformed into equivalent addition operations, resulting in \( Q \cdot K^T \in \mathbb{R} \). Moreover, because both \( Q \) and \( K \) are naturally non-negative, this preserves a non-negative attention map \( Q \cdot K^T \). Consequently, the mechanism eliminates the need for a softmax operation to enforce non-negativity within the attention map.
Furthermore, since \( K \) is a real-valued vector without boundaries, there is no necessity for a scaling factor \( s \) to manage large values resulting from the matrix multiplication, unlike the approaches in~\cite{zhou2022spikformer, zhou2024spikformer}. 
Additionally, given that \( V \) can also take values in \(\{-1, 0, 1\}\), the matrix product \( Q \cdot K^T \cdot V \) can similarly be expressed through addition operations. At the same time, unlike other matrix multiplication methods outlined in~\cite{zhou2022spikformer, zhou2024spikformer}, our framework allows for the result to include negative values, akin to what is seen in traditional ANNs.

In summary, the A$^2$OS$^2$A mechanism utilizes full-precision \( K \) and ternary spike \( V \) to enhance the representational capability of self-attention and minimize information loss. Additionally, all attention calculations are performed using addition operations, which align with the characteristics of SNNs.

%% file: sec/3_finalcopy.tex
\begin{table}[tp]
\caption{Ablation study for Spiking Transformer on CIFAR10/100. Param refers to the number of parameters. 
 Spiking Transformer-$L$-$D$ is a Spiking Transformer model with $L$ Spiking Transformer encoder blocks and $D$ feature embedding dimensions. }
\begin{center}\begin{adjustbox}{max width=1.02\linewidth} 
\begin{tabular}{ccccccc}
\toprule
  \multicolumn{1}{c}{\bf Methods} &\multicolumn{1}{c}{\bf \tabincell{c}{Param\\ (M)}}
&\bf\tabincell{c}{Time\\Step} &\bf\tabincell{c}{CIFAR10\\Acc} &\bf\tabincell{c}{CIFAR100\\Acc}\\
\midrule
    \multicolumn{1}{c}{\multirow{1}{*}{{Baseline-2-256}}}&2.59 & 4 & 94.39 & 76.00\\
    \multicolumn{1}{c}{\multirow{1}{*}{\textbf{Spiking Transformer-2-256}}}&2.59 & 4 & \textbf{94.91} & \textbf{76.96}\\
 \midrule   
    \multicolumn{1}{c}{\multirow{1}{*}{{Baseline-2-512}}}&10.23 & 4 & 95.51 & 78.83\\
    \multicolumn{1}{c}{\multirow{1}{*}{\textbf{Spiking Transformer-2-512}}}&10.23 & 4 & \textbf{96.42} & \textbf{79.90}\\
\bottomrule
\end{tabular}
\end{adjustbox}
\end{center}
\label{tab:ab}
\end{table}

\section{Experiment}
\label{sec:exp}
We evaluate our method on various datasets, including CIFAR-10/100~\cite{CIFAR-10} and ImageNet-1K~\cite{2009ImageNet}. The network architecture and experimental setup follow the baseline from Spike-driven Transformer~\cite{yao2024spike}. Further details regarding the experimental settings can be found in the Spike-driven Transformer~\cite{yao2024spike}.

\subsection{Ablation Study}
We conduct a series of ablation experiments to assess the effectiveness of the proposed Spiking Transformer, comparing it with the baseline model from~\cite{yao2024spike} on the CIFAR-10 and CIFAR-100 datasets. The results are summarized in Table~\ref{tab:ab}.

The baseline models used are Spike-driven Transformer-2-256 and Spike-driven Transformer-2-512. For the Transformer-2-256 with 4 timesteps, the baseline accuracy is 94.39\% on CIFAR-10 and 76.00\% on CIFAR-100, consistent with previous results. Our Spiking Transformer, using the same architecture, achieves notable improvements, with accuracies of 94.91\% and 76.96\% on CIFAR-10 and CIFAR-100, respectively—resulting in performance gains of approximately 0.5\% and 1.0\%.
For the larger model, Transformer-2-512 with 4 timesteps achieves baseline accuracies of 95.51\% and 78.43\% on CIFAR-10 and CIFAR-100. Our model with the same architecture shows substantial improvements, reaching accuracy scores of 96.42\% and 79.90\%, corresponding to gains of approximately 0.9\% and 1.1\% on CIFAR-10 and CIFAR-100, respectively.

\begin{table}[tp]
\caption{Comparison of the performance between Spiking Transformer and existing approaches on CIFAR10/100.}
\begin{center}\begin{adjustbox}{max width=1.02\linewidth} 
\begin{tabular}{ccccccc}
\toprule
  \multicolumn{1}{c}{\bf Methods} &\multicolumn{1}{c}{\bf \tabincell{c}{Param\\ (M)}}
&\bf\tabincell{c}{Time\\Step} &\bf\tabincell{c}{CIFAR10\\Acc} &\bf\tabincell{c}{CIFAR100\\Acc}\\
\midrule
    Hybrid training\cite{rathi2020enabling}  &9.27 &125 &92.22 &67.87\\
    Diet-SNN\cite{2020DIET}  &0.27 &10\textbf{/}5  & 92.54& 64.07\\
    STBP\cite{wu2018spatio}  &17.54&12 & 89.83&-\\
    STBP NeuNorm\cite{2018Direct}  &17.54 &12 &90.53& -\\
    TSSL-BP\cite{2020Temporal}  &17.54 &5 &91.41& -\\
    \multicolumn{1}{c}{\multirow{1}{*}{{STBP-tdBN\cite{2020Going}}}} &12.63 & 4 & 92.92 & 70.86\\
    TET\cite{deng2022temporal}   &12.63 & 4 & {94.44}& {74.47}\\
\midrule
    \multicolumn{1}{c}{\multirow{1}{*}{Spikformer-4-256}\cite{zhou2022spikformer}}&4.15 & 4 & 93.94 & {75.96}\\
    \multicolumn{1}{c}{\multirow{1}{*}{Spikformer-2-384}\cite{zhou2022spikformer}}&5.76 & 4 & {94.80} & {76.95}\\
    \multicolumn{1}{c}{\multirow{1}{*}{Spikformer-4-384}\cite{zhou2022spikformer}}&9.32 & 4 & {95.19} & {77.86}\\
\midrule
    \multicolumn{1}{c}{\multirow{1}{*}{Spikingformer-4-256}\cite{zhou2023spikingformer}}&4.15 & 4 & 94.77 & {77.43}\\
    \multicolumn{1}{c}{\multirow{1}{*}{Spikingformer-2-384}\cite{zhou2023spikingformer}}&5.76 & 4 & {95.22} & {78.34}\\
    \multicolumn{1}{c}{\multirow{1}{*}{Spikingformer-4-384}\cite{zhou2023spikingformer}}&9.32 & 4 & {95.61} & {79.09}\\
\midrule
    \multicolumn{1}{c}{\multirow{1}{*}{\textbf{Spiking Transformer-4-256}}}&4.15 & 4 & \textbf{94.96} & \textbf{77.49}\\
    \multicolumn{1}{c}{\multirow{1}{*}{\textbf{Spiking Transformer-2-384}}}&5.76 & 4 & \textbf{95.70} & \textbf{78.59}\\
    \multicolumn{1}{c}{\multirow{1}{*}{\textbf{Spiking Transformer-4-384}}}&9.32 & 4 & \textbf{96.32} & \textbf{79.69}\\
    \multicolumn{1}{c}{\multirow{1}{*}{\textbf{Spiking Transformer-2-512}}}&10.23 & 4 & \textbf{96.42} & \textbf{79.90}\\
\bottomrule
\end{tabular}
\end{adjustbox}
\end{center}
\label{tab:sd}
\end{table}

\subsection{CIFAR}

To further evaluate the performance of our method, we compare it with recent SNN-based Transformer approaches, including Spikformer~\cite{zhou2022spikformer}, Spikingformer~\cite{zhou2023spikingformer} and so on, using a range of models with varying embedding dimensions and numbers of transformer blocks on the CIFAR dataset. The CIFAR dataset consists of 50,000 training images and 10,000 test images, all with a resolution of $32 \times 32$. We maintain the experimental setup from \cite{zhou2022spikformer, zhou2023spikingformer, yao2024spike}, including network architecture, training configurations, etc. Table~\ref{tab:sd} presents the accuracy of our method compared to other models on the CIFAR dataset.
As shown in Table~\ref{tab:sd}, our Spiking Transformer-$2$-$384$ achieves an accuracy of $95.70\%$ on the CIFAR-10 dataset, surpassing TET ($94.44\%$) and even Spikingformer-4-384 ($94.97\%$). Notably, performance improves further with higher embedding dimensions and more transformer blocks. In particular, Spiking Transformer-4-384 shows a $1.13\%$ improvement over Spikformer-4-384 and a $0.71\%$ improvement over Spikingformer-4-384.
The performance gain of our Spiking Transformer is even more pronounced on more complex datasets such as CIFAR-100. Specifically, Spiking Transformer-4-384 achieves a notable improvement of $1.83\%$ over Spikformer-4-384.

\begin{table*}[tp]
\small
\caption{Comparison of the performance between Spiking Transformer and existing approaches on ImageNet-1k. }
\begin{center}
\resizebox{0.85\textwidth}{!}{
\begin{tabular}{cccccc}
\toprule
\multicolumn{1}{c}{\bf Category} 
&\multicolumn{1}{c}{\bf Methods} 
&\multicolumn{1}{c}{\bf Architecture}
&\multicolumn{1}{c}{\bf \tabincell{c}{Param\\ (M)}}
&\multicolumn{1}{c}{\bf\tabincell{c}{Time\\Step}}
&\multicolumn{1}{c}{\bf Acc}\\

 \midrule
 \multicolumn{1}{c}{\multirow{5}{*}{{ANN-to-SNN}}}
    &Hybrid training\cite{rathi2020enabling} &ResNet-34 &21.79 &250 &61.48\\
    & \multicolumn{1}{c}{\multirow{2}{*}{Spiking ResNet\cite{hu2021spiking}}} 
    &ResNet-34 &21.79  & 350 & 71.61 \\
    & &ResNet-50 &25.56 & 350 & 72.75 \\
    &QCFS\cite{bu2022optimal} &VGG-16 &138.42  &64 &72.85\\
    &COS\cite{hao2023bridging} &ResNet-34 &21.79  &8 &74.17\\
    
 \midrule

 \multicolumn{1}{c}{\multirow{8}{*}{{Directly Learning}}}
    &\multicolumn{1}{c}{\multirow{2}{*}{TET\cite{deng2022temporal}}} 
   & Spiking-ResNet-34 &21.79  &6 & {64.79} \\
   & & SEW-ResNet-34 &21.79  &4 & {68.00} \\
   & \multicolumn{1}{c}{\multirow{1}{*}{STBP-tdBN\cite{2020Going}}} &Spiking-ResNet-34 &21.79  & 6 & 63.72 \\
   & \multirow{4}{*}{SEW ResNet\cite{2021Deep}}  
    & SEW-ResNet-34 &21.79  &4 & 67.04 \\
   & & SEW-ResNet-50 &25.56   &4 & 67.78 \\
  &  & SEW-ResNet-101 &44.55   &4 & 68.76 \\
   & & SEW-ResNet-152 &60.19   &4 & 69.26 \\
   &Attention-SNN\cite{yao2023attention} &ResNet-104 &78.37  &4 &77.08\\
 \midrule
  \multicolumn{1}{c}{\multirow{5}{*}{{Directly Learning}}} &
   \multicolumn{1}{c}{\multirow{5}{*}{{Spikformer}\cite{zhou2022spikformer}}}  &\multicolumn{1}{c}{\multirow{1}{*}{Spikformer-8-384}} &16.81 &4 & 70.24\\
   & &\multicolumn{1}{c}{\multirow{1}{*}{Spikformer-6-512}}&23.37 & 4 & 72.46\\
   & &\multicolumn{1}{c}{\multirow{1}{*}{Spikformer-8-512}}&29.68 & 4 & 73.38\\
   & &\multicolumn{1}{c}{\multirow{1}{*}{Spikformer-10-512}}&36.01 & 4 & 73.68\\
   & &\multicolumn{1}{c}{\multirow{1}{*}{Spikformer-8-768}} &66.34 & 4 & 74.81\\
 \midrule
  \multicolumn{1}{c}{\multirow{3}{*}{{Directly Learning}}} &
   \multicolumn{1}{c}{\multirow{3}{*}{{Spikingformer}\cite{zhou2023spikingformer}}}  &\multicolumn{1}{c}{\multirow{1}{*}{Spikingformer-8-384}} &16.81 &4 & 72.45\\
   & &\multicolumn{1}{c}{\multirow{1}{*}{Spikingformer-8-512}}&29.68 & 4 & 74.79\\
   & &\multicolumn{1}{c}{\multirow{1}{*}{Spikingformer-8-768}} &66.34 & 4 & 75.85\\

 \midrule
  \multicolumn{1}{c}{\multirow{5}{*}{{Directly Learning}}} &
   \multicolumn{1}{c}{\multirow{5}{*}{{Spike-driven Transformer}\cite{yao2024spike}}}  &\multicolumn{1}{c}{\multirow{1}{*}{Spike-driven Transformer-8-384}} &16.81 &4 & 72.28\\
   & &\multicolumn{1}{c}{\multirow{1}{*}{Spike-driven Transformer-6-512}}&23.37 & 4 & 74.11\\
   & &\multicolumn{1}{c}{\multirow{1}{*}{Spike-driven Transformer-8-512}}&29.68 & 4 & 74.57\\
   & &\multicolumn{1}{c}{\multirow{1}{*}{Spike-driven Transformer-10-512}}&36.01 & 4 & 74.66\\
   & &\multicolumn{1}{c}{\multirow{1}{*}{Spike-driven Transformer-8-768}} &66.34 & 4 & 77.07\\
 \midrule
  \multicolumn{1}{c}{\multirow{4}{*}{{Directly Learning}}} &
   \multicolumn{1}{c}{\multirow{4}{*}{\textbf{Spiking Transformer}}}  &\multicolumn{1}{c}{\multirow{1}{*}{Spiking Transformer-8-384}} &16.81 &4 & \textbf{74.04}\\
   & &\multicolumn{1}{c}{\multirow{1}{*}{Spiking Transformer-6-512}}&23.37 & 4 & \textbf{76.22}\\
   & &\multicolumn{1}{c}{\multirow{1}{*}{Spiking Transformer-8-512}}&29.68 & 4 & \textbf{76.28}\\
   & &\multicolumn{1}{c}{\multirow{1}{*}{Spiking Transformer-10-512}}&36.01 & 4 & \textbf{78.66}\\
    \hline
\end{tabular}}
\end{center}
\label{tab:imagenet}
\end{table*}

\subsection{ImageNet} 
We further evaluated our approach on the challenging ImageNet-1K dataset. ImageNet-1K~\cite{2009ImageNet} is a widely-used benchmark for image classification, consisting of 1,000 categories. The dataset includes approximately 1.28 million training images and 50,000 test images. For both training and evaluation, the images are resized to a default resolution of 224 $\times$ 224 pixels.
To ensure consistency with previous work, we maintain the experimental setup from \cite{zhou2022spikformer, zhou2023spikingformer, yao2024spike}, including network architecture, training configurations, and other experimental details.

Table~\ref{tab:imagenet} presents a comparative analysis of recent SoTA methods evaluated on the ImageNet dataset. Notable models in this domain include Spikformer~\cite{zhou2022spikformer}, Spikingformer~\cite{zhou2023spikingformer}, and Spike-driven Transformer~\cite{yao2024spike}, which achieve accuracies of 73.38\%, 74.79\%, and 74.57\%, respectively, with Transformer-8-512 as the baseline configuration.
In comparison, our approach demonstrates a significant improvement, achieving an accuracy of 76.28\%, which represents a 1.49\% increase over the best-performing transformer-based SNN model, Spikingformer. This improvement highlights the effectiveness of our method in leveraging the potential of spiking neuron models within the transformer architecture, resulting in enhanced performance on large-scale datasets like ImageNet-1K.

Furthermore, even with a smaller network architecture, our Spiking Transformer-10-512 model outperforms the other methods, including those using Transformer-8-768 as the baseline. This demonstrates that our approach is not only more accurate but also more efficient, as it achieves superior results with fewer parameters, highlighting its scalability and robustness.


\section{Conclusion}
\label{sec:conclusion}

In this paper, we have introduced the Accurate Addition-Only Spiking Self-Attention (A$^2$OS$^2$A) mechanism as a novel method for integrating Spiking Neural Networks with Transformer models. Our approach tackles the inherent challenge of adapting the self-attention mechanism to the spiking paradigm by proposing a hybrid solution that combines binary, ReLU, and ternary spiking neurons. This innovative hybrid strategy enables the elimination of both softmax and scaling operations, which are traditionally essential in Transformer models, while preserving the fundamental advantages of addition-only computations in SNNs.

Through extensive experiments, we have demonstrated that A$^2$OS$^2$A outperforms existing SNN-based Transformer models in terms of both accuracy and efficiency. Specifically, our method achieves a SoTA accuracy of 78.66\% on the ImageNet-1K dataset. This performance, combined with the energy-efficient nature of SNNs, makes A$^2$OS$^2$A a promising candidate for real-world applications, especially those requiring high computational performance with low energy consumption.